\documentclass[10pt,twocolumn,letterpaper]{article}

\usepackage{wacv}              

\usepackage{algorithm}
\usepackage{algorithmic}
\usepackage{multirow}

\definecolor{wacvblue}{rgb}{0.21,0.49,0.74}
\usepackage[pagebackref,breaklinks,colorlinks,allcolors=wacvblue]{hyperref}

\def\wacvPaperID{1881} 
\def\confName{WACV}
\def\confYear{2026}

\title{Integrating Multi-scale and Multi-filtration Topological Features for \\Medical Image Classification\thanks{$\star$ indicates equal contribution. This research was supported in part by NSF Grants CCF-2523787, IIS-2334389, 2018900, BCS-2416846, and OAC-2417850, and DoD Grant W911NF2110169.}} 

\author{
Pengfei Gu$^{\star}$\\
The University of Texas Rio Grande Valley\\
{\tt\small pengfei.gu01@utrgv.edu}
\and
Huimin Li$^{\star}$\\
The University of Texas Rio Grande Valley\\
{\tt\small huimin.li01@utrgv.edu}
\and
Haoteng Tang\\
The University of Texas Rio Grande Valley\\
{\tt\small haoteng.tang@utrgv.edu}
\and 
Dongkuan (DK) Xu\\
North Carolina State University\\
{\tt\small dxu27@ncsu.edu}
\and
Erik Enriquez\\
The University of Texas Rio Grande Valley\\
{\tt\small erik.enriquez01@utrgv.edu}
\and 
DongChul Kim\\
The University of Texas Rio Grande Valley\\
{\tt\small dongchul.kim@utrgv.edu}
\and 
Bin Fu\\
The University of Texas Rio Grande Valley\\
{\tt\small bin.fu@utrgv.edu}
\and 
Danny Z. Chen\\
University of Notre Dame\\
{\tt\small dchen@nd.edu}
}

\begin{document}
\maketitle
\begin{abstract}
Modern deep neural networks have shown remarkable performance in medical image classification. However, such networks either emphasize pixel-intensity features instead of fundamental anatomical structures (e.g., those encoded by topological invariants), or they capture only simple topological features via single-parameter persistence.
In this paper, we propose a new topology-guided classification framework that extracts multi-scale and multi-filtration persistent topological features and integrates them into vision classification backbones.
For an input image, we first compute cubical persistence diagrams (PDs) across multiple image resolutions/scales. We then develop a ``vineyard'' algorithm that consolidates these PDs into a single, stable diagram capturing signatures at varying granularities, from global anatomy to subtle local irregularities that may indicate early-stage disease.
To further exploit richer topological representations produced by multiple filtrations, we design a cross-attention-based neural network that directly processes the consolidated final PDs. The resulting topological embeddings are fused with feature maps from CNNs or Transformers. By integrating multi-scale and multi-filtration topologies into an end-to-end architecture, our approach enhances the model's capacity to recognize complex anatomical structures.
Evaluations on three public datasets show consistent, considerable improvements over strong baselines and state-of-the-art methods, demonstrating the value of our comprehensive topological perspective for robust and interpretable medical image classification.

\end{abstract}
    
\section{Introduction}
\label{sec:intro}
Deep neural networks (DNNs) have become commonly-used methods for medical image analysis (e.g., classification), powered by expressive visual representations learned by convolutional networks (CNNs)~\cite{alexnet,vggnet,googlnet,resnet,densenet,hu2018squeeze,gu2021k,gu2024boosting} and, more recently, by Vision Transformers (ViTs)~\cite{dosovitskiy2020image,liu2021swin,liu2022swin} that model long-range dependencies via self-attention. 
Despite these advances, mainstream architectures rely predominantly on encoding intensity-driven appearance and local texture features, while \emph{global, robust anatomical structures} --- such as connected components and loops that characterize tissues, organs, and lesions --- remain weakly represented. 
This gap is consequential to medical practice, where object topology often conveys clinically meaningful patterns (e.g., lesion cavities, branches, and enclosure)~\cite{gu2025topoimages,adame2025topo,gu2024self,stucki2023topologically,wang2025topotxr}.

Topological data analysis (TDA)~\cite{topo_intro}, and in particular \emph{persistent homology} (PH)~\cite{edelsbrunner2002topological}, offers a powerful approach to quantify such structures. 
PH tracks the birth and death of topological features along a filtration, summarizing them as a multi-set of points in a \emph{persistence diagram} (PD). 
However, efforts of leveraging PDs in deep learning (DL) methods are non-trivial.
{(1) Established vectorization methods:} Previous methods such as persistent images~\cite{pers_image}, persistent landscapes~\cite{pers_land}, PersLay~\cite{perslay}, PLLay~\cite{pllay}, and projection layers~\cite{hofer2017deep,hofer2020graph} convert a PD (a variable-size multi-set) into a fixed-length vector via a mathematically prescribed encoding. These processes are often arduous and task-specific (e.g., choices of kernels, grids, and smoothing), and can suffer information bottlenecks that discard fine-grained geometry or higher-order interactions among PD points. 
{(2) Data-driven PD encoding:} More recently, PHG-Net~\cite{peng2024phg} learns a permutation-invariant encoder that directly consumes PD points and co-trains with a vision backbone, thereby reducing hand-crafted design and retaining task-relevant signals. Yet, PHG-Net, like most prior work, operates on \emph{single-parameter} persistence from a single filtration, and does not explicitly target \emph{stable multi-scale} or \emph{multi-filtration} topological signatures.

\begin{figure*}[t!]
    \centering
    \includegraphics[width=0.85\textwidth]{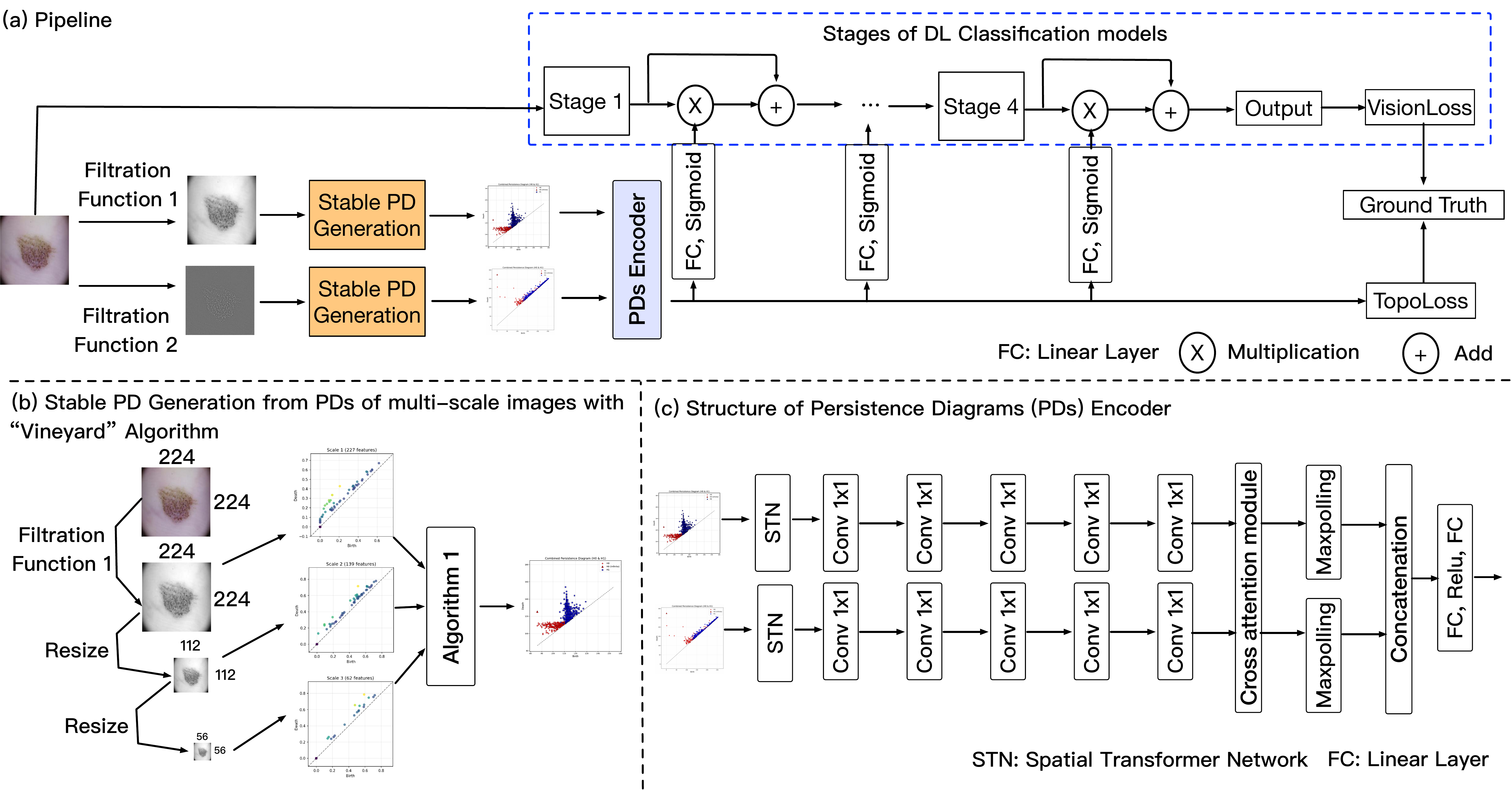}
    \caption{(a) Illustrating the pipeline of our proposed approach. 
    (b) Illustrating the generation of a stable persistence diagram from the multi-scale persistence diagrams using the ``vineyard'' tracking Algorithm~\ref{alg:vineyard}  (using a single filtration function as an example). 
    (c) The architecture of the proposed persistence diagrams (PDs) encoder.
    VisionLoss and TopoLoss represent the vision loss and topological loss, respectively. We use an image from the ISIC 2018 dataset and a CNN backbone for illustration. For simplicity, the batch normalization and rectified linear unit that follow each 1D convolution layer are omitted.
    } 
    \label{fig:pipeline}
\end{figure*}

We hypothesize that exploiting stable, complementary topological cues can enhance medical image classification by aligning learned representations with clinically meaningful structures across scales and filtrations. Hence, in this paper, we propose a new approach that (1) extracts \emph{stable} multi-scale, multi-filtration topological signatures and (2) injects them throughout the backbone to improve classification performance.
Specifically, given an image, we compute cubical PDs at multiple resolutions (scales) under multiple filtration functions (e.g., intensity and gradient magnitude). We then introduce a ``vineyard'' tracking algorithm that associates homologous points across scales and selects reliable trajectories, producing a \emph{stable} PD that captures topological signatures ranging from organ-level shapes to subtle boundary irregularities.
To further leverage multi-filtration topological features, we generate stable PDs under different filtrations and introduce a \emph{PDs encoder} to fuse complementary cues across filtrations via bidirectional cross-attention. The resulting topological embedding is then injected into the backbone at \emph{multiple layers} via a channel-wise gating module with residual connections, enabling topology-aware refinement of intermediate features in the backbone models.

We evaluate the proposed method with standard DL classification networks on three public medical image datasets. Extensive experiments show that our method yields significant improvements over state-of-the-art (SOTA) methods, 
demonstrating the effectiveness of our new approach.

Our contributions are three-fold: (1) We propose a new framework for extracting stable multi-scale, multi-filtration topological features for medical image classification.
(2) Our method is model-agnostic and can incorporate topological features into any vision backbone, including CNNs and Transformers.
(3) We achieve considerable performance improvements on three public datasets compared to SOTA medical image classification methods.

\section{Method}
\label{sec:method}
Fig.~\ref{fig:pipeline}(a) shows the overall pipeline of our approach.
Generally, given an input image $I$, we first compute PDs from multi-scale versions of $I$ using different filtration functions. Stable PDs are then generated from these PDs using Algorithm~\ref{alg:vineyard}. A PDs encoder is applied to transform the stable PDs into a topological feature vector, which is integrated with the feature maps produced by the backbone DL classification model at each stage to enable feature refinement.

This section begins with a brief review of how PDs are constructed from images. We then describe the ``vineyard'' algorithm for generating stable PDs from multi-scale images. Next, we present the design of our PDs encoder for embedding stable PDs into topological feature vectors. Finally, we discuss how these vectors are integrated to guide the DL classification model in feature refinement for medical image classification.

\begin{algorithm}[t!]
\caption{``Vineyard'' Tracking Algorithm for a Stable Persistence Diagram}
\label{alg:vineyard}
\begin{algorithmic}[1]
\REQUIRE Ordered persistence diagrams (PDs) $\{PD_1,\ldots,PD_n\}$, matching threshold $\tau_m$, stability threshold $\tau_s$, distance metric $\mathsf{dist}$
\ENSURE Stable PD $PD_{\mathrm{stable}}$
\STATE $V_{\mathrm{active}} \leftarrow \emptyset$ \COMMENT{map: current-point index $\mapsto$ vine}
\STATE $V_{\mathrm{complete}} \leftarrow \emptyset$
\FOR{$i=1$ \TO $n-1$}
  \STATE $A \leftarrow PD_i[:,(b,d)]$, $B \leftarrow PD_{i+1}[:,(b,d)]$
  \STATE Compute pairwise distance matrix $D$ between points of $A$ and $B$ using $\mathsf{dist}$
  \STATE Compute optimal matching assignment $M$ by the Hungarian algorithm on $D$
  \STATE $V_{\mathrm{next}} \leftarrow \emptyset$ ; $S_{\mathrm{curr}} \leftarrow \emptyset$ ; $S_{\mathrm{next}} \leftarrow \emptyset$
  \FOR{each matched pair $(a\in A,\, b\in B)$ in $M$}
     \IF{$D[a,b] \le \tau_m$}
        \STATE Form segment $s=(i,\;a{\to}b,\;\text{distance}=D[a,b],\;\text{pers}_{a},\text{pers}_{b})$
        \IF{$a \in V_{\mathrm{active}}$}
            \STATE Append $s$ to vine $V_{\mathrm{active}}[a]$; \quad $V_{\mathrm{next}}[b]\!\leftarrow\! V_{\mathrm{active}}[a]$
        \ELSE
            \STATE Create new vine $v$ with segment $s$; \quad $V_{\mathrm{next}}[b]\!\leftarrow\! v$
        \ENDIF
        \STATE $S_{\mathrm{curr}} \leftarrow S_{\mathrm{curr}}\cup\{a\}$; \quad $S_{\mathrm{next}} \leftarrow S_{\mathrm{next}}\cup\{b\}$
     \ENDIF
  \ENDFOR
  \FOR{each $a$ such that $a\in V_{\mathrm{active}}$ and $a\notin S_{\mathrm{curr}}$}
     \STATE Set $\text{death\_scale}$ of $V_{\mathrm{active}}[a]$ to $i{+}1$ and move it to $V_{\mathrm{complete}}$
  \ENDFOR
  \STATE $V_{\mathrm{active}} \leftarrow V_{\mathrm{next}}$
\ENDFOR
\STATE Move all vines remaining in $V_{\mathrm{active}}$ to $V_{\mathrm{complete}}$ with $\text{death\_scale}=n$
\STATE $PD_{\mathrm{stable}} \leftarrow \emptyset$
\FOR{each vine $v \in V_{\mathrm{complete}}$}
   \STATE Compute stability score $\sigma(v)$
   \IF{$\sigma(v) \ge \tau_s$}
      \STATE Let $m \leftarrow \lfloor |v.\text{segments}|/2 \rfloor$ (middle segment index)
      \STATE Add the middle segment's source point $(b,d)$ to $PD_{\mathrm{stable}}$
   \ENDIF
\ENDFOR
\RETURN $PD_{\mathrm{stable}}$
\end{algorithmic}
\end{algorithm}

\subsection{Constructing Persistence Diagrams of Images}
\label{subsec:ph_pd}

Given an input image $I$, we model its pixel grid as a \emph{cubical complex} $C$ that takes each pixel as a $0$-cell (vertex) and encodes neighborhood connectivity through $1$-cells (edges), $2$-cells (squares), and higher-dimensional cubes when appropriate \cite{wagner2011efficient,kaji2020cubical}. 
Let $f:I\rightarrow\mathbb{R}$ be a \emph{filtration function} (e.g., the raw intensity or a derived scalar field). 
Varying a threshold $\tau$ induces a nested family of super-level subcomplexes
$
C^{(\tau)} \;:=\; \{\, x\in I \mid f(x)\ge \tau \,\},
$
and, as $\tau$ decreases through finitely many critical values $\tau_1 \ge \tau_2 \ge \cdots \ge \tau_m$, we obtain the \emph{super-level set filtration}
$
\emptyset \;\subseteq\; C^{(\tau_1)} \;\subseteq\; C^{(\tau_2)} \;\subseteq\; \cdots \;\subseteq\; C^{(\tau_m)} \;=\; I.
$

{PH} \cite{edelsbrunner2002topological} tracks how homology classes evolve along this filtration. 
For 2D images, we mainly consider $k\in\{0,1\}$, corresponding to connected components ($0$-D) and cycles/loops ($1$-D) in homology classes. 
A topological feature that first \emph{appears} (is born) at threshold value $b$ and then \emph{disappears/merges} (dies) at threshold value $d< b$ is encoded by a pair $(b,d)$. 
Its \emph{persistence (lifespan)} is
$
pers \;=\; b - d \;> 0,
$
quantifying the range of scales over which the feature remains topologically salient (longer lifespans usually indicate more meaningful structures). PH provides a principled multi-scale description of image topology.

\begin{figure*}[t!]
    \centering
    \includegraphics[width=0.78\textwidth]{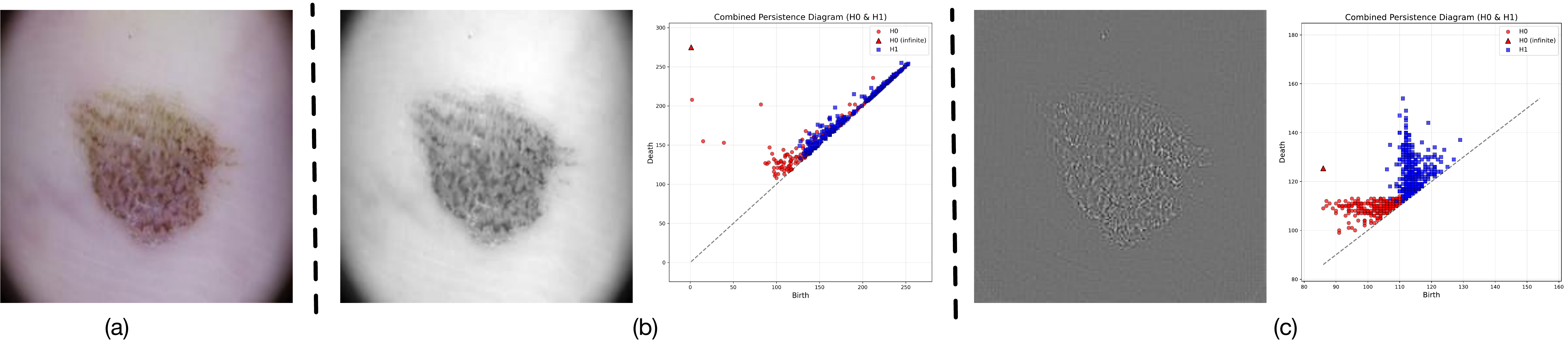}
    \caption{Examples of persistence diagrams in an image from the ISIC 2018 dataset under different filtrations. (a) An original image; (b) image with intensity values and its corresponding persistence diagram; (c) image with gradient magnitude values and its corresponding persistence diagram. In the PDs, red circles denote 0-D persistent homology (H0), and blue squares denote 1-D persistent homology (H1).}
    \label{fig:pds}
\end{figure*}

Collecting all birth--death pairs in a fixed dimension $k$ yields the \emph{$k$-th PD},
$
D_I^{k} \;=\; \big\{\, (b_i^{(k)}, d_i^{(k)}) \in \mathbb{R}^2 \;\big|\; d_i^{(k)} < b_i^{(k)} \, \}, 
k\in\{0,1\},
$
a multi-set of points that compactly summarizes when features occur and how long they persist across the filtration. PDs serve as stable, interpretable shape descriptors that encode the births and deaths of connected components and loops across thresholds.
For visualization, one often plots points with $b<d$ by switching to an equivalent sub-level convention (or by replacing $f$ with $-f$). This monotone reparameterization keeps the underlying topology and persistences unchanged.
Fig.~\ref{fig:pds} shows some examples of PDs in images from the ISIC 2018 dataset~\cite{codella2019skin} using the intensity filtration function (i.e., magnitude values of the three RGB channels of pixels) and gradient filtration function (i.e., average gradient magnitude values computed by the Laplacian operator across the three RGB channels).

\subsection{Computing Stable Multi-scale Topological Features with the ``Vineyard'' Algorithm}
\label{subsec:algorithm} 
A single-scale topological analysis risks being either too coarse --- missing subtle, localized cues such as early-stage lesions or boundary irregularities --- or too fine, thereby overlooking global organ-level structures and anatomical context. By computing PDs across multiple resolutions, $\{PD_1,\dots,PD_n\}$, we capture topological signatures at varying granularities: from large-scale anatomical shapes to fine-scale locally irregular patterns that may be clinically meaningful. Hence, this multi-scale representation is better aligned with the heterogeneous nature of medical imagery, where pathology can manifest at multiple spatial scales simultaneously.

However, not all multi-scale features are equally reliable for downstream tasks (e.g., classification). Some features are transient (noise or resolution artifacts), while others persist consistently across scales and thus reflect stable anatomical or pathological structures. 
To identify and retain the most reliable topological features, we propose to explicitly track putative correspondences of PD points across adjacent scales (forming ``vines''), and then select those vines whose trajectories are both smooth (small inter-scale movement) and topologically significant (high persistence).

Given ordered PDs across the scales, we conduct a ``vineyard'' tracking procedure to assemble trajectories and select stable features. Its main steps are as follows.
\textbf{(1) Multi-Scale Diagrams:}
Load PDs at $n$ image resolutions $\{PD_i\}_{i=1}^n$, where each $PD_i$ contains birth--death pairs $(b,d)$ for a fixed homology degree.
\textbf{(2) Persistence-Aware Costs:}
For each consecutive pair $(PD_i,PD_{i+1})$, we define a distance between points $p=(b_p,d_p)\in PD_i$ and $q=(b_q,d_q)\in PD_{i+1}$.
\textbf{(3) Bipartite Matching:}
Construct the pairwise distance matrix $D$ on $(PD_i,PD_{i+1})$, and compute an optimal matching assignment with the Hungarian algorithm~\cite{Network-Flows}. Keep only matches with $D[a,b]\le \tau_m$ (the matching threshold) to avoid implausible jumps.
\textbf{(4) Vine Assembly:}
Each accepted match creates a segment from $p\in PD_i$ to $q\in PD_{i+1}$, annotated with its distance and endpoint persistences. If the source point is already tracked, we \emph{extend} its vine; otherwise, we \emph{start} a new vine at scale $i$. Tracked points not matched at the next scale are \emph{terminated} (the vine ``dies''). 
\textbf{(5) Stability Scoring:}
For a vine $v$ with segments $s$, we compute a persistence-weighted stability
$
\sigma(v)=\frac{\sum_{s\in v} w_s \cdot \frac{1}{1+\mathrm{dist}(s)}}{\sum_{s\in v} w_s},
\qquad
w_s=\max\!\Big(0.1,\ \tfrac{\mathrm{pers}_{\text{from}}(s)+\mathrm{pers}_{\text{to}}(s)}{2}\Big),
$
which rewards smooth trajectories while emphasizing high-persistence structure. We keep vines with $\sigma(v)\ge \tau_s$.
\textbf{(6) Stable Diagram Construction:}
Each retained vine is summarized by a representative birth--death pair taken at its \emph{medial} scale (middle segment in practice), yielding the final stable PD $PD_{\mathrm{stable}}$ to be used as a compact, reliable descriptor for downstream learning.

\begin{figure*}[t!]
    \centering
    \includegraphics[width=0.78\textwidth]{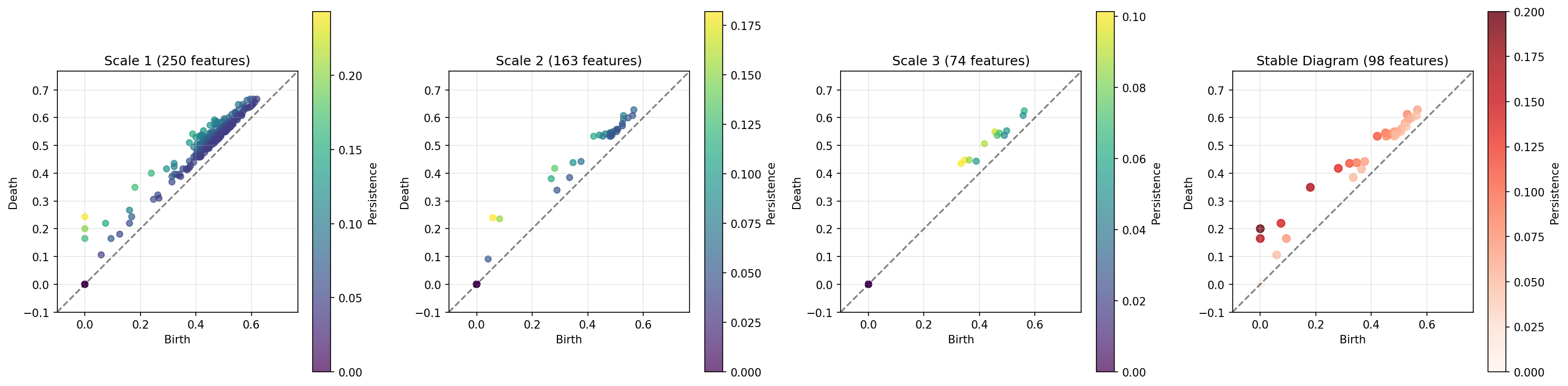}
    \caption{Examples of stable persistence diagrams generated from multi-scale persistence diagrams in images of the Kvasir dataset. Here, scale 1 corresponds to $224 \times 224$, scale 2 to $112 \times 112$, and scale 3 to $56 \times 56$.}
    \label{fig:multi-pds}
\end{figure*}

Algorithm~\ref{alg:vineyard} gives the procedure for computing the final stable PD for one homology degree (e.g., $H_0$, $H_1$, or $H_2$), and the final stable diagram is formed by a degree-wise union:
$
PD_{\mathrm{stable}}
\;=\;
\bigcup_{r \in \{0,1,2\}} PD^{(r)}_{\mathrm{stable}}.
$
In our experiments, we compute the PDs at three image resolutions (i.e., $224\times 224$, $112\times112$, and $56\times 56$) to generate the final stable PD (see examples in Fig.~\ref{fig:multi-pds}). 
Unlike~\cite{cohen2006vines}, which studied vineyards along a continuous family of filter functions, our approach constructs ``vineyards'' by matching PDs computed at different image resolutions.

\subsection{Exploiting Multi\texorpdfstring{-}{-}filtration Topological Features with the Persistence Diagrams Encoder}
\label{subsec:mf_pd_encoder}

A single filtration (e.g., raw pixel intensity) captures only one projection of image structures. 
Applying \emph{multiple} filtrations to the same image, such as intensity and gradient magnitude, accentuates complementary aspects of anatomy~\cite{gu2025topoimages}. One filtration may sharpen tissue interfaces, while another may emphasize gradual transitions or pathology-related gradients. 
Formally, for given filtration functions $\{f^{(t)}: I\!\to\!\mathbb{R}\}_{t=1}^T$ on image $I$, each $f^{(t)}$ induces a stable PD using Algorithm~\ref{alg:vineyard}, providing complementary multi-scale and multi-filtration topological features.  

PDs are the most prevalent descriptors in PH, and numerous handcrafted mappings (e.g., persistence images, landscapes, Betti curves) transform PDs into vectors for downstream learning. 
However, these mappings impose fixed functional forms and may discard task-relevant information. 
To avoid such information drawbacks, we propose a \emph{learning}, end-to-end PDs encoder that directly encodes PDs. 

First, for each stable PD, we encode the sets of 2D points in the PD into expressive vectors. A naive approach for this would aggregate $\{(b,d)\}$ by max-/avg-pooling; but such pooling is order invariant yet \emph{interaction-poor}, risking loss of higher-order relations.
Therefore, we first \emph{lift} each 2D point to a higher-dimensional feature space that enables point-point interactions before applying a symmetric aggregation. Specifically, we adapt a PointNet style~\cite{pointnet} Spatial Transformer Network (STN)~\cite{jaderberg2015spatial} to learn spatial transformations that make our network invariant to rigid transformations of 3D point clouds. Given an input point cloud represented as a tensor of shape $(\text{channels}, \text{points})$, where the channels correspond to coordinate dimensions (typically 3 for $xyz$ coordinates), the STN predicts a single $3 \times 3$ transformation matrix.
The STN architecture consists of a feature extraction backbone followed by a regression head. The backbone uses three 1D convolutional layers with channel dimensions $\{64, 128, 1024\}$, each followed by batch normalization and ReLU activation:
$
x^{(1)} = \text{ReLU}(\text{BN}(\text{Conv1d}_{c \rightarrow 64}(x^{(0)})))$,
$
x^{(2)} = \text{ReLU}(\text{BN}(\text{Conv1d}_{64 \rightarrow 128}(x^{(1)})))$,
$
x^{(3)} = \text{ReLU}(\text{BN}(\text{Conv1d}_{128 \rightarrow 1024}(x^{(2)}))),
$
where $x^{(0)}$ is the input tensor and $c$ is the number of input channels. To achieve permutation invariance across the points, we apply symmetric max-pooling,
$f = \max_{i} x^{(3)}_i$,
yielding a global feature vector $f \in \mathbb{R}^{1024}$. This global feature is then passed through a three-layer MLP regression head with dimensions $\{512, 256, 9\}$:
$
h^{(1)} = \text{ReLU}(\text{BN}(\text{Linear}_{1024 \rightarrow 512}(f)))$,
$
h^{(2)} = \text{ReLU}(\text{BN}(\text{Linear}_{512 \rightarrow 256}(h^{(1)})))$,
$
\hat{T} = \text{Linear}_{256 \rightarrow 9}(h^{(2)}).
$
To ensure that the transformation matrix is initialized as an identity and remains close to it during early training, we add an identity matrix residual connection,
$T = \hat{T} + I_{3 \times 3}$,
where $I_{3 \times 3}$ is the $3 \times 3$ identity matrix flattened to a 9-dimensional vector and then reshaped back to matrix form. The final transformation matrix $T \in \mathbb{R}^{3 \times 3}$ is applied to the input points via matrix multiplication, enabling the network to learn the invariance to rotations, translations, and uniform scaling of the input point cloud.

Then, we process the vectors through multiple 1D convolutions with batch normalization and ReLU activations, and fuse complementary information across the filtrations. Stable PDs produced by different filtrations contain non-redundant cues (e.g., a loop persistent in the gradient-filtration but not in intensity). 
An effective encoder must discover correspondences and complementarities across the PD sets and fuse them without collapsing distinct signals. 
To capture correspondences among different feature sets, we employ a multi-head cross-attention mechanism. Given query features $Q \in \mathbb{R}^{N_q \times D}$, key features $K \in \mathbb{R}^{N_k \times D}$, and value features $V \in \mathbb{R}^{N_v \times D}$, we compute attended features through scaled dot-product attention.
The input features are first projected through learned linear transformations,
$Q' = QW_Q, K' = KW_K,$ and $V'$ $=$ $VW_V,$
where $W_Q, W_K, W_V \in \mathbb{R}^{D \times D}$ are learnable parameter matrices.
For multi-head attention with $H$ heads, we reshape the projected features into head-specific representations with dimension $d = D/H$:
$Q_h = Q'[:, h \cdot d:(h+1) \cdot d], K_h = K'[:, h \cdot d:(h+1) \cdot d],$ and $V_h = V'[:, h \cdot d:(h+1) \cdot d].$
The attention mechanism computes similarity scores between queries and keys, applies softmax normalization, and aggregates values:
$\text{Attention}_h(Q, K, V) = \text{softmax}\left(\frac{Q_h K_h^T}{\sqrt{d}}\right) V_h.$
The outputs from all attention heads are concatenated and projected through a final linear layer,
$\text{MultiHead}(Q, K, V) = \text{Concat}(\text{Attention}_1, \ldots, \text{Attention}_H) W_O$,
where $W_O \in \mathbb{R}^{D \times D}$ is the output projection matrix.
This cross-attention mechanism enables selective information exchange among different feature sets, allowing the model to identify and leverage correspondences across heterogeneous inputs. The multi-head structure captures diverse types of relationships simultaneously, while the scaled attention prevents gradient vanishing in high-dimensional feature spaces.
Finally, we concatenate the feature vectors and pass it to two MLP layers to produce the final topological feature vector, capturing richer topological representations. Fig.~\ref{fig:pipeline}(c) illustrates the details of the PDs encoder that we design.

Following~\cite{gu2025topoimages}, we use intensity and gradient filtration functions in our experiments.

\subsection{Integrating Topological Features in DL Classification Models}
\label{subsec:refineing}
A straightforward integration strategy is to concatenate the topological embedding $t\!\in\!\mathbb{R}^{M}$ (from the PDs encoder) with the penultimate deep features and train a joint classifier. 
However, this \emph{late fusion} has two drawbacks:
(i) gradients for the topology and image branches are largely computed in isolation during back-propagation, limiting cross-branch interaction; 
(ii) concatenation at the final stage prevents topology from shaping intermediate, multi-scale representations where semantic and anatomical cues emerge.

To address this issue, we employ an attention-based residual gating method~\cite{peng2024phg}.
Let $\mathcal{L}$ index a set of intermediate layers of a CNN or Transformer backbone. 
For a layer $\ell\!\in\!\mathcal{L}$, denote the features as
$F^{(\ell)}\!\in\!\mathbb{R}^{H_\ell\times W_\ell\times C_\ell}$ for CNNs 
(or $F^{(\ell)}\!\in\!\mathbb{R}^{N_\ell\times C_\ell}$ for token features). 
We inject a topology-conditioned \emph{channel gate} $\mathbf{g}^{(\ell)}\!\in\!\mathbb{R}^{C_\ell}$ computed from $t$:
$
\mathbf{g}^{(\ell)} \;=\; \sigma\!\Big(\, W^{(\ell)}_{1}\, \mathrm{ReLU}\!\big( W^{(\ell)}_{2}\, t \big) \Big),
\label{eq:chan-gate}
$
where $W^{(\ell)}_{2}\!\in\!\mathbb{R}^{\frac{C_\ell}{r}\times M}$ reduces $t$ to a bottleneck of width $C_\ell/r$, 
$W^{(\ell)}_{1}\!\in\!\mathbb{R}^{C_\ell\times \frac{C_\ell}{r}}$ restores channel dimension, 
$\sigma$ denotes sigmoid, and $r\!\ge\!1$ balances accuracy and cost.
The gate modulates features multiplicatively (broadcast over spatial positions or tokens):
$
\hat{F}^{(\ell)} \;=\; F^{(\ell)} \odot \mathbf{g}^{(\ell)} .
\label{eq:mul-gate}
$
To preserve identity mapping and stabilize training, we adopt a residual update with a learnable scalar $\alpha_\ell$:
$
F^{(\ell)}_{\text{ref}} \;=\; F^{(\ell)} \;+\; \alpha_\ell \,\hat{F}^{(\ell)} ,
\quad \alpha_\ell \ge 0.
$
We optimize the network end-to-end with the following combination losses:
$
\mathcal{J}
\;=\;
\mathcal{L}_{\mathrm{cls}}^{\mathrm{vis}}\!\left(\hat{y}^{\mathrm{vis}},\, y\right)
\;+\;
\lambda\, \mathcal{L}_{\mathrm{cls}}^{\mathrm{topo}}\!\left(\hat{y}^{\mathrm{topo}},\, y\right),
$
where $\mathcal{L}_{\mathrm{cls}}^{\mathrm{vis}}$ and $\mathcal{L}_{\mathrm{cls}}^{\mathrm{topo}}$ are cross-entropy terms for the vision backbone and topological branch respectively, 
$\hat{y}^{\mathrm{vis}}$ and $\hat{y}^{\mathrm{topo}}$ are the class-score outputs of the two branches, $y$ is ground truth, and $\lambda$ is sets as $0.1$ to control the contribution of the topology-guided supervision (see Fig.~\ref{fig:pipeline}(a) for more details).

\section{Experiments}\label{sec:exp}
\subsection{Datasets}

{\bf ISIC 2018~\cite{codella2019skin}:} 
This skin lesion classification dataset contains 10,015 training images and 193 validation images across 7 classes of skin disease lesions. For fair comparison, we follow the dataset split strategy in~\cite{zhuang2018skin} using five-fold cross validation.

{\bf Kvasir~\cite{pogorelov2017kvasir}:} 
This dataset contains 4,000 images, annotated and verified by experienced endoscopists, including 8 classes showing anatomical landmarks, phatological findings or endoscopic procedures in
the GI tract, with hundreds of images for each class. Following~\cite{yue2024medmamba}, we split the dataset into 2,408/392/1,200 images for training/validation/test, and apply five-fold cross validation.

{\bf CBIS-DDSM~\cite{lee2017curated}:} 
The Curated Breast Imaging Subset of the Digital Database for Screening Mammography (CBIS-DDSM) consists of 1,566 participants and 6,775 studies, categorized as benign or malignant. We adopt the train/test split provided by the dataset, with 20\% of the cases reserved for testing and the rest for training.

\subsection{Implementation Details}\label{imp}
Our experiments are conducted with the PyTorch framework. We employ the Cubical Ripser software~\cite{kaji2020cubical} to compute PDs.
Model training is performed on an NVIDIA Tesla V100 Graphics Card with 32GB GPU memory using the AdamW optimizer~\cite{loshchilov2017decoupled} with a weight decay = $0.005$. 
The learning rate is 0.0001, and the number of training epochs is 400 for the experiments. The batch size for each case is set as the maximum size allowed by the GPU.
We apply standard data augmentation, such as rotation, random flip, random crop, etc. to avoid overfitting.

\begin{table*}[ht]

\centering

\caption{Experimental results of different methods on the ISIC 2018 dataset. The best results are highlighted in \textbf{bold}.
}


\label{main_isic}

\begin{tabular}{l|c|c|c|c}\hline
Method &  Accuracy & AUC & Sensitivity & Specificity \\
\hline
		
ResNet152~\cite{resnet} &87.68	&96.70	&70.82	&96.60 \\

ResNet152 + PHG~\cite{peng2024phg} & 89.63	&97.47	&76.62	&97.07 \\

ResNet152 + PHG + multi-scale& 90.16	&98.34	&78.97	&97.31 \\

ResNet152 + PHG + multi-scale + multi-filtrations (Ours) & 91.56	&98.81	&81.11	&97.48 \\\hline

SENet154~\cite{hu2018squeeze} &88.03	&97.03	&75.94	&96.83 \\

SENet154 + PHG~\cite{peng2024phg} & 90.47	&97.75	&78.55	&97.31 \\

SENet154 + PHG + multi-scale& 91.13	&98.20	&81.13	&97.43 \\

SENet154 + PHG + multi-scale + multi-filtrations (Ours)& 92.58	&98.95	&82.73	&97.82 \\\hline

SwinV2-B~\cite{liu2022swin} & 89.03	&97.77	&79.63	&97.03 \\ 

SwinV2-B + PHG~\cite{peng2024phg} & 90.87	&98.13	&82.36	&97.39 \\

SwinV2-B + PHG + multi-scale&  91.60	&98.74	&83.97	&97.68 \\

SwinV2-B + PHG + multi-scale + multi-filtrations (Ours)& \textbf{92.94}	&\textbf{99.10}	&\textbf{85.56}	&\textbf{98.12} \\\hline
        
\end{tabular}	

\end{table*}

\begin{table*}[t!]
\centering

\caption{Experimental results of different methods on the Kvasir dataset. The best results are highlighted in \textbf{bold}.
} 

\label{main_kvasir}
\begin{tabular}{l|c|c|c|c}\hline

Method &  Accuracy & AUC & Sensitivity & Specificity \\

\hline
  
ResNet152~\cite{resnet} & 75.92	&96.33	&75.92	&96.99 \\

ResNet152 + PHG~\cite{peng2024phg} & 77.52	&97.03	&77.52	&97.59 \\

ResNet152 + PHG + multi-scale& 78.27	&97.18	&78.27	&97.70 \\

ResNet152 + PHG + multi-scale + multi-filtrations (Ours)& 79.52	&97.63	&79.52	&\textbf{97.88} \\\hline

SENet154~\cite{hu2018squeeze} & 78.00	&97.45	&78.00	&96.50 \\

SENet154 + PHG~\cite{peng2024phg} & 79.00	&97.85	&79.00	&97.00 \\

SENet154 + PHG + multi-scale& 79.33	&98.01	&79.33	&97.19 \\

SENet154 + PHG + multi-scale + multi-filtrations (Ours)& 80.42	&\textbf{98.11}	&80.42	&97.43 \\\hline

SwinV2-B~\cite{liu2022swin} & 78.58	&97.35	&78.58	&97.08 \\ 

SwinV2-B + PHG~\cite{peng2024phg} & 79.78	&97.05	&79.78	&97.09\\

SwinV2-B + PHG + multi-scale& 80.36	&97.15	&80.36	&97.27 \\

SwinV2-B + PHG + multi-scale + multi-filtrations (Ours)& \textbf{81.52}	&97.25	&\textbf{81.52}	&97.40 \\\hline
        
\end{tabular}	

\end{table*}

\begin{table*}[t!]
\centering

\caption{Experimental results of different methods on the CBIS-DDSM dataset. The best results are highlighted in \textbf{bold}.
}

\label{main_cbis}

\begin{tabular}{l|c|c|c|c}\hline

Method &  Accuracy & AUC & Sensitivity & Specificity \\

\hline

ResNet152~\cite{resnet} & 72.16	&77.81	&73.72	&70.93 \\

ResNet152 + PHG~\cite{peng2024phg} & 74.29	&79.16	&72.24	&71.96 \\

ResNet152 + PHG + multi-scale& 75.14	&79.86	&74.05	&73.35 \\

ResNet152 + PHG + multi-scale + multi-filtrations (Ours)& 76.43	&80.29	&75.19	&74.47 \\\hline

SENet154~\cite{hu2018squeeze} & 70.88	& 77.44	& 72.74	& 70.70 \\

SENet154 + PHG~\cite{peng2024phg} & 74.15	&80.23	&73.44	&72.94 \\

SENet154 + PHG + multi-scale& 74.95	&80.95	&75.07	&73.31 \\

SENet154 + PHG + multi-scale + multi-filtrations (Ours)& 77.23	& 82.43	& 76.80	& 74.91 \\\hline

SwinV2-B~\cite{liu2022swin} & 73.15	&80.12	&72.27	&73.22 \\ 

SwinV2-B + PHG~\cite{peng2024phg} & 75.74	&83.98	&75.13	&75.03 \\

SwinV2-B + PHG + multi-scale& 76.28	&85.97	&76.84	&76.35 \\

SwinV2-B + PHG + multi-scale + multi-filtrations (Ours)& \textbf{78.40}	&\textbf{87.22}	&\textbf{78.35}	&\textbf{77.89} \\\hline
        
		
	\end{tabular}
\end{table*}

\begin{table}[t!]
\centering
\caption{Accuracy results for different matching and stability thresholds in Algorithm~\ref{alg:vineyard}, evaluated on the ISIC 2018 dataset with ResNet152.}
\label{tab:patch-size}
\centering
 \scalebox{0.83}{%
\begin{tabular}{l|c|c|c}
\hline
\multirow{3}{*}{Matching Threshold $\tau_m$}  & \multicolumn{3}{c}{Stability Threshold $\tau_s$}  \\
 &{\begin{tabular}[c]{@{}l@{}} 0.6\end{tabular}} & \begin{tabular}[c]{@{}l@{}} 0.7\end{tabular} & \begin{tabular}[c]{@{}l@{}} 0.8\end{tabular} \\
\hline
0.2  & 89.00 &90.67 & 88.83  \\\hline
0.3   &90.50  & \textbf{91.56} & 90.17  \\\hline
0.4   &89.83  &90.33 &  89.58 \\\hline
\end{tabular}	
}
\end{table}

\subsection{Results and Analysis}
\label{subsec:results}

We evaluate three strong vision backbones: ResNet152~\cite{resnet}, SENet154~\cite{hu2018squeeze}, and SwinV2-B~\cite{liu2022swin}. 
For each backbone, we conduct a controlled ablation by progressively enabling:
(i) a topology branch following PHG~\cite{peng2024phg}, 
(ii) \emph{+ multi-scale} vineyard stabilization across image resolutions, and 
(iii) \emph{+ multi-filtrations} (full model) which fuses intensity- and gradient-derived stable PDs via our PDs encoder and applies multi-scale topology-guided refinement (Sec.~\ref{sec:method}). 
We report results in Accuracy, AUC, Sensitivity, and Specificity.

We compare our method against PHG-Net~\cite{peng2024phg}, a SOTA topology-based DL classification method. 
Note that we do not additionally compare to two common PD vectorization approaches, PersLay~\cite{perslay} and PLLay~\cite{pllay}, because PHG-Net has already shown that it outperforms these vectorizers on medical image classification tasks. 
We also do not compare to TopoImages~\cite{gu2025topoimages}, as it is a hand-crafted data augmentation method that encodes local topology rather than learning from PDs for medical image classification. 
Besides, we do not include CuMPerLay~\cite{nuwagira2025cumperlay}, a differentiable layer for cubical multiparameter persistence, since our method builds on \emph{stable single-parameter PDs} aggregated across scales and filtrations rather than full multiparameter persistence.

Tables~\ref{main_isic},~\ref{main_kvasir}, and~\ref{main_cbis} present the experimental results. We observe the following. (1) Our method consistently outperforms all three backbones and PHG-Net~\cite{peng2024phg} across the three datasets and the performance metrics. In particular, we improve the accuracy by 4.55\% over SENet154 on the ISIC 2018 dataset and by 3.08\% over PHG-Net~\cite{peng2024phg} on the CBIS-DDSM dataset.
These results show the effectiveness of the proposed method.
(2) Adding the ``vineyard'' algorithm across resolutions (while using a single filtration) yields average gains of 
$+0.64$ Acc (range $[+0.33,+0.85]$), 
$+0.64$ AUC ($[+0.10,+1.99]$), 
$+1.48$ Sens ($[+0.33,+2.58]$), and 
$+0.47$ Spec ($[+0.11,+1.39]$) 
when averaged over all the datasets/backbones. 
This supports the premise that \emph{scale\mbox{-}stable} topological signatures suppress resolution\mbox{-}specific artifacts while retaining clinically salient structures.
(3) Fusing stable PDs from different filtrations (intensity and gradient) provides the largest improving boost. For example, on the CBIS-DDSM dataset, the accuracy is improved by 2.28\% and 2.12\% using SENet154 and SwinV2-B, respectively. In addition,
improvements are particularly pronounced in Sensitivity (crucial for screening) while Specificity remains high, indicating that topology\mbox{-}guided refinement improves true\mbox{-}positive detection without sacrificing false\mbox{-}positive control.
Together, these parts yield consistent Accuracy/AUC gains and marked Sensitivity improvements across the backbones/datasets, validating the effectiveness of multi\mbox{-}scale, multi\mbox{-}filtration topologies for medical image classification.

\subsection{Ablation Study}
\label{subsec:ablation-thresholds}
The ``vineyard'' algorithm (Algorithm~\ref{alg:vineyard}) has two key hyperparameters: the \emph{matching threshold} $\tau_m$, which controls how aggressively PD points are linked across adjacent scales, and the \emph{stability threshold} $\tau_s$, which filters completed vines based on their persistence-weighted smoothness score $\sigma(v)$.
Intuitively, a small $\tau_m$ risks breaking valid correspondences (vine fragmentation), while a large $\tau_m$ may link unrelated points (spurious vines). Similarly, a small $\tau_s$ retains noisy or unstable vines, whereas a large $\tau_s$ may prune away informative but moderately stable structures.

Our experiments evaluate $\tau_m \in \{0.2,0.3,0.4\}$ and $\tau_s \in \{0.6,0.7,0.8\}$ on the ISIC 2018 dataset using ResNet152, with all the other settings fixed.
Table~\ref{tab:patch-size} reports the accuracy results. The highest accuracy, 91.56\%, occurs at $(\tau_m,\tau_s)=(0.3,0.7)$.
Therefore, we set $\tau_m=0.3$ and $\tau_s=0.7$ as the default values and adopt this setting in our experiments, as it balances recall (sufficient cross-scale linking) with precision (filtering to stable vines).

Furthermore, we conduct experiments to investigate the effects of different distance metrics in the ``vineyard'' algorithm. We compare three metrics between points $p=(b_p,d_p)$ and $q=(b_q,d_q)$ with persistence $\mathrm{pers}_p=\max(0,d_p-b_p)$: 
(1) Wasserstein Distance~\cite{mileyko2011probability}: $d_{\text{W}}(p,q)=\sqrt{(b_p-b_q)^2+(d_p-d_q)^2}$, 
(2) Persistence-scaled Distance: $d_{\text{PS}}(p,q)=d_{\text{W}}(p,q)\,\Big/\!\Big(1+\tfrac{\mathrm{pers}_p+\mathrm{pers}_q}{2}\Big)$, 
and (3) Relative Persistence Distance: $d_{\text{RP}}(p,q)=d_{\text{W}}(p,q)\,\Big(1+\tfrac{|\mathrm{pers}_p-\mathrm{pers}_q|}{\max(\mathrm{pers}_p,\mathrm{pers}_q)}\Big)$. 
We evaluate on the CBIS\mbox{-}DDSM dataset with SwinV2\mbox{-}B.
Table~\ref{tab:fusion-meth} presents the accuracy results. We observe that the Relative Persistence Distance attains the best performance ({78.40}\%), outperforming 
Wasserstein (73.30\%; $\uparrow5.10\%$) and 
Persistence-Scaled (73.72\%; $\uparrow4.68\%$).
Therefore, we use the Relative Persistence Distance ({$d_{\text{RP}}$}) as the default distance metric in Algorithm~\ref{alg:vineyard} for our experiments.

\begin{table}[t]
    \centering
    \caption{Results of different distance metrics in Algorithm~\ref{alg:vineyard}, evaluated on the CBIS-DDSM dataset with SwinV2-B.}
    \label{tab:fusion-meth}
    \scalebox{0.83}{%
    \begin{tabular}{l|l}
        \hline
        Method &Accuracy \\\hline
        Wasserstein Distance&	73.30	\\
       \hline
      Persistence-Scaled Distance
          &	73.72	\\\hline
         Relative Persistence Distance
          &	\textbf{78.40} 	\\
       \hline
    \end{tabular}
}
\end{table}

\subsection{Time and Parameter Complexity}
\label{subsec:time-param}
Our pipeline for computing stable PDs consists of the following main steps:  
(i) constructing cubical PDs at three image scales ($224 \times 224$, $112 \times 112$, and $56 \times 56$) with two filtrations (intensity and gradient), and  
(ii) applying the ``vineyard'' algorithm across the PDs of adjacent scales to compute stable PDs.  
Here, we report the execution time on the Kvasir dataset (4,000 images). For the intensity filtration, computing PDs takes 32 seconds, 28 seconds, and 20 seconds at the $224 \times 224$, $112 \times 112$, and $56 \times 56$ resolutions, respectively. The ``vineyard'' algorithm requires 22 minutes and 56 seconds to generate stable PDs from these PDs. For the gradient filtration, computing PDs takes 38 seconds, 30 seconds, and 18 seconds at the three scales, respectively, and the ``vineyard'' algorithm takes 1 hour, 13 minutes, and 8 seconds for computing the stable PD.  
It is worth noting that PD computation is performed offline, and this process needs to be conducted only once. 

To examine the parameter complexity, Table~\ref{tab:paras} reports the numbers of parameters for SwinV2-B, SwinV2-B + PHG, and our proposed approach for comparison.

\begin{table}[t]
    \centering
   \caption{Parameter complexity comparison of different methods.}
    \label{tab:paras}
    \scalebox{0.83}{%
    \begin{tabular}{l|l}
        \hline
        Method &  Number of Parameters \\\hline
        SwinV2-B&	86.908 M	\\
       \hline
      SwinV2-B + PHG
          &	94.455 M	\\\hline
         Ours
          &	144.238 M 	\\
       \hline
    \end{tabular}
}
\end{table}


\section{Conclusions}
In this paper, we introduced a new topology-guided framework for medical image classification that augments intensity-based representations with stable topological signatures which are aggregated across scales and filtrations.
We computed cubical PDs at multiple resolutions/scales and consolidate them via a ``vineyard'' algorithm to capture stable topological signatures ranging from global anatomy to fine local irregularities. 
To further exploit richer topological representations derived from multiple filtrations, we designed a cross-attention-based neural network specifically for processing the final PDs, producing task-relevant embeddings that are fused with CNN/Transformer features in an end-to-end pipeline. 
Experiments on three public datasets showed that our method yields consistent, considerable gains over the baselines and SOTA methods, demonstrating the effectiveness of our new approach. 
{
    \small
    \bibliographystyle{ieeenat_fullname}
    \bibliography{main}

\begin{thebibliography}{39}
\providecommand{\natexlab}[1]{#1}
\providecommand{\url}[1]{\texttt{#1}}
\expandafter\ifx\csname urlstyle\endcsname\relax
  \providecommand{\doi}[1]{doi: #1}\else
  \providecommand{\doi}{doi: \begingroup \urlstyle{rm}\Url}\fi

\bibitem[Adame et~al.(2025)Adame, Nunez, Vazquez, Gurrola, Li, Tang, Fu, and Gu]{adame2025topo}
Diego Adame, Jose~A Nunez, Fabian Vazquez, Nayeli Gurrola, Huimin Li, Haoteng Tang, Bin Fu, and Pengfei Gu.
\newblock {Topo-VM-UNetV2}: Encoding topology into {Vision Mamba UNet} for polyp segmentation.
\newblock In \emph{Proceedings of the IEEE 38th International Symposium on Computer-Based Medical Systems}, pages 258--263, 2025.

\bibitem[Adams et~al.(2017)Adams, Emerson, Kirby, Neville, Peterson, Shipman, Chepushtanova, Hanson, Motta, and Ziegelmeier]{pers_image}
Henry Adams, Tegan Emerson, Michael Kirby, Rachel Neville, Chris Peterson, Patrick Shipman, Sofya Chepushtanova, Eric Hanson, Francis Motta, and Lori Ziegelmeier.
\newblock {Persistence Images}: A stable vector representation of persistent homology.
\newblock \emph{Journal of Machine Learning Research}, 18\penalty0 (8):\penalty0 1--35, 2017.

\bibitem[Ahuja et~al.(1993)Ahuja, Magnanti, and Orlin]{Network-Flows}
Ravindra~K. Ahuja, Thomas~L. Magnanti, and James~B. Orlin.
\newblock \emph{Network Flows: Theory, Algorithms, and Applications}.
\newblock Prentice Hall, 1993.

\bibitem[Bubenik et~al.(2015)]{pers_land}
Peter Bubenik et~al.
\newblock Statistical topological data analysis using persistence landscapes.
\newblock \emph{Journal of Machine Learning Research}, 16\penalty0 (1):\penalty0 77--102, 2015.

\bibitem[Carri{\`e}re et~al.(2020)Carri{\`e}re, Chazal, Ike, Lacombe, Royer, and Umeda]{perslay}
Mathieu Carri{\`e}re, Fr{\'e}d{\'e}ric Chazal, Yuichi Ike, Th{\'e}o Lacombe, Martin Royer, and Yuhei Umeda.
\newblock {PersLay}: A neural network layer for persistence diagrams and new graph topological signatures.
\newblock In \emph{Proceedings of the International Conference on Artificial Intelligence and Statistics}, pages 2786--2796, 2020.

\bibitem[Codella et~al.(2019)Codella, Rotemberg, Tschandl, Celebi, Dusza, Gutman, Helba, Kalloo, Liopyris, Marchetti, et~al.]{codella2019skin}
Noel Codella, Veronica Rotemberg, Philipp Tschandl, M~Emre Celebi, Stephen Dusza, David Gutman, Brian Helba, Aadi Kalloo, Konstantinos Liopyris, Michael Marchetti, et~al.
\newblock Skin lesion analysis toward melanoma detection 2018: A challenge hosted by the international skin imaging collaboration ({ISIC}).
\newblock \emph{arXiv preprint arXiv:1902.03368}, 2019.

\bibitem[Cohen-Steiner et~al.(2006)Cohen-Steiner, Edelsbrunner, and Morozov]{cohen2006vines}
David Cohen-Steiner, Herbert Edelsbrunner, and Dmitriy Morozov.
\newblock Vines and vineyards by updating persistence in linear time.
\newblock In \emph{Proceedings of the Twenty-second Annual Symposium on Computational Geometry}, pages 119--126, 2006.

\bibitem[Dosovitskiy et~al.(2020)Dosovitskiy, Beyer, Kolesnikov, Weissenborn, Zhai, Unterthiner, Dehghani, Minderer, Heigold, Gelly, et~al.]{dosovitskiy2020image}
Alexey Dosovitskiy, Lucas Beyer, Alexander Kolesnikov, Dirk Weissenborn, Xiaohua Zhai, Thomas Unterthiner, Mostafa Dehghani, Matthias Minderer, Georg Heigold, Sylvain Gelly, et~al.
\newblock An image is worth 16x16 words: Transformers for image recognition at scale.
\newblock \emph{arXiv preprint arXiv:2010.11929}, 2020.

\bibitem[Edelsbrunner and Harer(2010)]{topo_intro}
Herbert Edelsbrunner and John Harer.
\newblock \emph{Computational Topology: An Introduction}.
\newblock American Mathematical Soc, 2010.

\bibitem[Edelsbrunner et~al.(2002)Edelsbrunner, Letscher, and Zomorodian]{edelsbrunner2002topological}
Herbert Edelsbrunner, David Letscher, and Afra Zomorodian.
\newblock Topological persistence and simplification.
\newblock \emph{Discrete \& Computational Geometry}, 28:\penalty0 511--533, 2002.

\bibitem[Gu et~al.(2021)Gu, Zheng, Zhang, Wang, and Chen]{gu2021k}
Pengfei Gu, Hao Zheng, Yizhe Zhang, Chaoli Wang, and Danny~Z Chen.
\newblock {kCBAC-Net}: Deeply supervised complete bipartite networks with asymmetric convolutions for medical image segmentation.
\newblock In \emph{Proceedings of the International Conference on Medical Image Computing and Computer-Assisted Intervention}, pages 337--347, 2021.

\bibitem[Gu et~al.(2024{\natexlab{a}})Gu, Zhang, Li, Wang, and Chen]{gu2024self}
Pengfei Gu, Yejia Zhang, Huimin Li, Chaoli Wang, and Danny~Z Chen.
\newblock Self pre-training with topology-and spatiality-aware masked autoencoders for {3D} medical image segmentation.
\newblock \emph{arXiv preprint arXiv:2406.10519}, 2024{\natexlab{a}}.

\bibitem[Gu et~al.(2024{\natexlab{b}})Gu, Zhao, Wang, Peng, Zhang, Sapkota, Wang, and Chen]{gu2024boosting}
Pengfei Gu, Zihan Zhao, Hongxiao Wang, Yaopeng Peng, Yizhe Zhang, Nishchal Sapkota, Chaoli Wang, and Danny~Z Chen.
\newblock Boosting medical image classification with segmentation foundation model.
\newblock \emph{arXiv preprint arXiv:2406.11026}, 2024{\natexlab{b}}.

\bibitem[Gu et~al.(2025)Gu, Wang, Zhang, Li, Wang, and Chen]{gu2025topoimages}
Pengfei Gu, Hongxiao Wang, Yejia Zhang, Huimin Li, Chaoli Wang, and Danny Chen.
\newblock {TopoImages}: Incorporating local topology encoding into deep learning models for medical image classification.
\newblock \emph{arXiv preprint arXiv:2508.01574}, 2025.

\bibitem[He et~al.(2016)He, Zhang, Ren, and Sun]{resnet}
Kaiming He, Xiangyu Zhang, Shaoqing Ren, and Jian Sun.
\newblock Deep residual learning for image recognition.
\newblock In \emph{Proceedings of the IEEE Conference on Computer Vision and Pattern Recognition}, pages 770--778, 2016.

\bibitem[Hofer et~al.(2017)Hofer, Kwitt, Niethammer, and Uhl]{hofer2017deep}
Christoph Hofer, Roland Kwitt, Marc Niethammer, and Andreas Uhl.
\newblock Deep learning with topological signatures.
\newblock \emph{Advances in Neural Information Processing Systems}, 30, 2017.

\bibitem[Hofer et~al.(2020)Hofer, Graf, Rieck, Niethammer, and Kwitt]{hofer2020graph}
Christoph Hofer, Florian Graf, Bastian Rieck, Marc Niethammer, and Roland Kwitt.
\newblock Graph filtration learning.
\newblock In \emph{Proceedings of the International Conference on Machine Learning}, pages 4314--4323, 2020.

\bibitem[Hu et~al.(2018)Hu, Shen, and Sun]{hu2018squeeze}
Jie Hu, Li Shen, and Gang Sun.
\newblock Squeeze-and-excitation networks.
\newblock In \emph{Proceedings of the IEEE Conference on Computer Vision and Pattern Recognition}, pages 7132--7141, 2018.

\bibitem[Huang et~al.(2017)Huang, Liu, Van Der~Maaten, and Weinberger]{densenet}
Gao Huang, Zhuang Liu, Laurens Van Der~Maaten, and Kilian~Q Weinberger.
\newblock Densely connected convolutional networks.
\newblock In \emph{Proceedings of the IEEE Conference on Computer Vision and Pattern Recognition}, pages 4700--4708, 2017.

\bibitem[Jaderberg et~al.(2015)Jaderberg, Simonyan, Zisserman, et~al.]{jaderberg2015spatial}
Max Jaderberg, Karen Simonyan, Andrew Zisserman, et~al.
\newblock Spatial {Transformer} networks.
\newblock \emph{Advances in Neural Information Processing Systems}, 28, 2015.

\bibitem[Kaji et~al.(2020)Kaji, Sudo, and Ahara]{kaji2020cubical}
Shizuo Kaji, Takeki Sudo, and Kazushi Ahara.
\newblock {Cubical Ripser}: Software for computing persistent homology of image and volume data.
\newblock \emph{arXiv preprint arXiv:2005.12692}, 2020.

\bibitem[Kim et~al.(2020)Kim, Kim, Zaheer, Kim, Chazal, and Wasserman]{pllay}
Kwangho Kim, Jisu Kim, Manzil Zaheer, Joon~Sik Kim, Fr{\'e}d{\'e}ric Chazal, and Larry Wasserman.
\newblock {PLLay}: Efficient topological layer based on persistence landscapes.
\newblock \emph{Advances in Neural Information Processing Systems}, 2020.

\bibitem[Krizhevsky et~al.(2017)Krizhevsky, Sutskever, and Hinton]{alexnet}
Alex Krizhevsky, Ilya Sutskever, and Geoffrey~E Hinton.
\newblock {ImageNet} classification with deep convolutional neural networks.
\newblock \emph{Communications of the ACM}, 60\penalty0 (6):\penalty0 84--90, 2017.

\bibitem[Lee et~al.(2017)Lee, Gimenez, Hoogi, Miyake, Gorovoy, and Rubin]{lee2017curated}
Rebecca~Sawyer Lee, Francisco Gimenez, Assaf Hoogi, Kanae~Kawai Miyake, Mia Gorovoy, and Daniel~L Rubin.
\newblock A curated mammography data set for use in computer-aided detection and diagnosis research.
\newblock \emph{Scientific Data}, 4\penalty0 (1):\penalty0 1--9, 2017.

\bibitem[Liu et~al.(2021)Liu, Lin, Cao, Hu, Wei, Zhang, Lin, and Guo]{liu2021swin}
Ze Liu, Yutong Lin, Yue Cao, Han Hu, Yixuan Wei, Zheng Zhang, Stephen Lin, and Baining Guo.
\newblock {Swin Transformer}: Hierarchical vision {Transformer} using shifted windows.
\newblock In \emph{Proceedings of the IEEE/CVF International Conference on Computer Vision}, pages 10012--10022, 2021.

\bibitem[Liu et~al.(2022)Liu, Hu, Lin, Yao, Xie, Wei, Ning, Cao, Zhang, Dong, et~al.]{liu2022swin}
Ze Liu, Han Hu, Yutong Lin, Zhuliang Yao, Zhenda Xie, Yixuan Wei, Jia Ning, Yue Cao, Zheng Zhang, Li Dong, et~al.
\newblock Swin {Transformer} v2: Scaling up capacity and resolution.
\newblock In \emph{Proceedings of the IEEE/CVF Conference on Computer Vision and Pattern Recognition}, pages 12009--12019, 2022.

\bibitem[Loshchilov and Hutter(2017)]{loshchilov2017decoupled}
Ilya Loshchilov and Frank Hutter.
\newblock Decoupled weight decay regularization.
\newblock \emph{arXiv preprint arXiv:1711.05101}, 2017.

\bibitem[Mileyko et~al.(2011)Mileyko, Mukherjee, and Harer]{mileyko2011probability}
Yuriy Mileyko, Sayan Mukherjee, and John Harer.
\newblock Probability measures on the space of persistence diagrams.
\newblock \emph{Inverse Problems}, 27\penalty0 (12):\penalty0 124007, 2011.

\bibitem[Nuwagira et~al.()Nuwagira, Korkmaz, Coskunuzer, and Birdal]{nuwagira2025cumperlay}
Brighton Nuwagira, Caner Korkmaz, Baris Coskunuzer, and Tolga Birdal.
\newblock {CuMPerLay}: Learning cubical multiparameter persistence vectorizations.
\newblock In \emph{Proceedings of the 2025 Fall Southeastern Sectional Meeting}.

\bibitem[Peng et~al.(2024)Peng, Wang, Sonka, and Chen]{peng2024phg}
Yaopeng Peng, Hongxiao Wang, Milan Sonka, and Danny~Z Chen.
\newblock {PHG-Net}: Persistent homology guided medical image classification.
\newblock In \emph{Proceedings of the IEEE/CVF Winter Conference on Applications of Computer Vision}, pages 7583--7592, 2024.

\bibitem[Pogorelov et~al.(2017)Pogorelov, Randel, Griwodz, Eskeland, de~Lange, Johansen, Spampinato, Dang-Nguyen, Lux, Schmidt, et~al.]{pogorelov2017kvasir}
Konstantin Pogorelov, Kristin~Ranheim Randel, Carsten Griwodz, Sigrun~Losada Eskeland, Thomas de Lange, Dag Johansen, Concetto Spampinato, Duc-Tien Dang-Nguyen, Mathias Lux, Peter~Thelin Schmidt, et~al.
\newblock Kvasir: A multi-class image dataset for computer aided gastrointestinal disease detection.
\newblock In \emph{Proceedings of the 8th ACM on Multimedia Systems Conference}, pages 164--169, 2017.

\bibitem[Qi et~al.(2017)Qi, Su, Mo, and Guibas]{pointnet}
Charles~R Qi, Hao Su, Kaichun Mo, and Leonidas~J Guibas.
\newblock {PointNet}: Deep learning on point sets for {3D} classification and segmentation.
\newblock In \emph{Proceedings of the IEEE Conference on Computer Vision and Pattern Recognition}, pages 652--660, 2017.

\bibitem[Simonyan and Zisserman(2014)]{vggnet}
Karen Simonyan and Andrew Zisserman.
\newblock Very deep convolutional networks for large-scale image recognition.
\newblock \emph{arXiv preprint arXiv:1409.1556}, 2014.

\bibitem[Stucki et~al.(2023)Stucki, Paetzold, Shit, Menze, and Bauer]{stucki2023topologically}
Nico Stucki, Johannes~C Paetzold, Suprosanna Shit, Bjoern Menze, and Ulrich Bauer.
\newblock Topologically faithful image segmentation via induced matching of persistence barcodes.
\newblock In \emph{Proceedings of the International Conference on Machine Learning}, pages 32698--32727, 2023.

\bibitem[Szegedy et~al.(2015)Szegedy, Liu, Jia, Sermanet, Reed, Anguelov, Erhan, Vanhoucke, and Rabinovich]{googlnet}
Christian Szegedy, Wei Liu, Yangqing Jia, Pierre Sermanet, Scott Reed, Dragomir Anguelov, Dumitru Erhan, Vincent Vanhoucke, and Andrew Rabinovich.
\newblock Going deeper with convolutions.
\newblock In \emph{Proceedings of the IEEE Conference on Computer Vision and Pattern Recognition}, pages 1--9, 2015.

\bibitem[Wagner et~al.(2011)Wagner, Chen, and Vu{\c{c}}ini]{wagner2011efficient}
Hubert Wagner, Chao Chen, and Erald Vu{\c{c}}ini.
\newblock Efficient computation of persistent homology for cubical data.
\newblock In \emph{Proceedings of the Topological Methods in Data Analysis and Visualization II: Theory, Algorithms, and Applications}, pages 91--106, 2011.

\bibitem[Wang et~al.(2025)Wang, Zou, Sakla, Partyka, Rawal, Singh, Zhao, Ling, Huang, Prasanna, et~al.]{wang2025topotxr}
Fan Wang, Zhilin Zou, Nicole Sakla, Luke Partyka, Nil Rawal, Gagandeep Singh, Wei Zhao, Haibin Ling, Chuan Huang, Prateek Prasanna, et~al.
\newblock {TopoTxR}: A topology-guided deep convolutional network for breast parenchyma learning on {DCE-MRIs}.
\newblock \emph{Medical Image Analysis}, 99:\penalty0 103373, 2025.

\bibitem[Yue and Li(2024)]{yue2024medmamba}
Yubiao Yue and Zhenzhang Li.
\newblock {MedMamba}: Vision {Mamba} for medical image classification.
\newblock \emph{arXiv preprint arXiv:2403.03849}, 2024.

\bibitem[Zhuang et~al.(2018)Zhuang, Li, Manivannan, Wang, Zhang, Liu, Pan, Jiang, and Yin]{zhuang2018skin}
Jiaxin Zhuang, Weipeng Li, Siyamalan Manivannan, Roy Wang, JianGuo Zhang, Jihan Liu, Jiahui Pan, Gongfa Jiang, and Ziyu Yin.
\newblock Skin lesion analysis towards melanoma detection using deep neural network ensemble.
\newblock \emph{ISIC Challenge}, 2018\penalty0 (2):\penalty0 1--6, 2018.

\end{thebibliography}
}

\end{document}